\documentclass{article}
\usepackage{amsmath,amsfonts,amssymb}
\usepackage{verbatim}
\usepackage{graphicx}
\usepackage{tabularx}
\usepackage{booktabs}
\usepackage{url}
\usepackage{epstopdf}
\usepackage{subfigure}
\usepackage{paralist}
\usepackage{grffile}
\usepackage{textcomp}

\newcommand{\e}{\mathbb{E}}
\renewcommand{\v}{\mathbb{V}\text{ar}}

\newcommand{\n}{\mathcal{N}}
\renewcommand{\r}{\mathbb{R}}

\newcommand{\di}{\!^{^*}}

\newcommand{\tdot}{\!\cdot\!}
\newcommand{\dupcolval}{\textquotestraightdblbase}

\title{Moment based estimation
of stochastic Kronecker graph parameters}
\author{
David F. Gleich\\
Sandia National Laboratories\thanks{Sandia National Laboratories is a multi-program laboratory
       managed and operated by Sandia Corporation, a wholly owned
       subsidiary of Lockheed Martin Corporation, for the U.S.
       Department of Energy's National Nuclear Security Administration
       under contract DE-AC04-94AL85000.}\\
\and
Art B. Owen\\
Stanford University
}
\date{June 2011}
\begin{document}

\maketitle

\begin{abstract}
Stochastic Kronecker graphs supply a parsimonious
model for large sparse real world graphs.
They can specify the distribution of a large random
graph using only three or four parameters.  Those parameters
have however proved difficult to choose in specific applications.
This article looks at method of moments estimators
that are computationally much simpler than maximum likelihood.
The estimators are fast and in our
examples, they typically yield Kronecker parameters
with expected feature counts closer to a given
graph than we get from KronFit. The improvement was
especially prominent for the number of triangles
in the graph.
\end{abstract}

\section{Introduction}

Stochastic Kronecker graphs were introduced by 
\cite{lesk:chak:klei:falo:2005} as a method
for simulating very large random graphs. 
Random synthetic graphs are used to test graph algorithms
and to understand observed properties of graphs.  
By using simulated graphs, instead of real measured ones,
it is possible to test algorithms on graphs larger or denser
than  presently observed ones.
Simulated graphs also allow one to judge which features of a real graph
are likely to hold in other similar graphs and which are idiosyncratic
to the given data.

Stochastic Kronecker graphs are able to serve these purposes
through a model that has only three or four parameters.
Parameter estimation poses unique challenges
for those graphs.  
The main problem is that for a graph
with $N$ nodes, the likelihood 
has contributions from $N!$ permutations of the
nodes \cite{lesk:falo:2007}.
In practice, many thousands or millions of randomly sampled
permutations are used to estimate the likelihood.
Even then it takes more
than $O(N^2)$ work to evaluate the likelihood contribution
from one of the permutations.

In this paper we present a method of moments
strategy for parameter estimation. While moment
methods can be inefficient compared to maximum
likelihood, statistical efficiency is of reduced
importance for enormous samples and in settings
where the dominant error is lack of fit.
The method equates expected to observed counts
for edges, triangles, hairpins ($2$-stars or wedges)
and tripins ($3$-stars).
The Kronecker model gives quite tractable formulas
for these moments.

The outline of this paper is as follows.
Section~\ref{sec:definitions} defines Kronecker
graphs and introduces some notation.
Section~\ref{sec:moment} derives the expected
feature counts.
Section~\ref{sec:solving} describes how to solve
method of moment equations for the parameters
of the Kronecker graph model.
Section~\ref{sec:examples} presents some examples
on fitting Kronecker models to some real
world graphs. We compare several moment based
ways to estimate Kronecker graph parameters
and find the most reliable results come from a criterion
that sums squared relative errors between observed
and expected features. 
We find that the fitted Kronecker models usually 
underestimate the number of triangles
compared to the real graphs.
While our parameter estimates underestimate
triangle counts and some other features, 
we find that they provide much closer matches
than some previously published parameters fit
by KronFit.
Section~\ref{sec:simu} fits parameters to graphs
that were randomly generated from the Kronecker model.
We find that the estimated parameters closely
track their generating values, with some
small bias when a parameter is at the extreme
range of valid values. 
Section~\ref{sec:conclusions} has our conclusions.

The data for our examples is online at
$$\text{\url{https://dgleich.com/gitweb/?p=kgmoments;a=summary}}$$
along with the code used to estimate Kronecker
parameters.

\section{The Kronecker model}\label{sec:definitions}

Given a node set $\n$ of cardinality $N\ge 1$,  
and a matrix $P_{ij}\in[0,1]$ defined
over $i,j\in\n$, a random graph $G^*(P)$ is one where
the edge $[ij]$ exists with probability $P_{ij}$
and all $N^2$ edges exist or don't independently.
The graph $G^*$ includes loops $[ii]$ and may possibly
include both $[ij]$ and $[ji]$.
We snip these out by defining
the random graph $G(P)$ with edges
$[ij]$ only when $i\ne j$ and
$[\max(i,j),\min(i,j)]\in G^*$,
using any non-random ordering of $\n$.
Both $G^*$ and $G$ are in fact probability weighted 
ensembles of graphs,
but for simplicity we describe them as single random graphs.
We assume that $P$ is a symmetric matrix and so
the ordering of nodes does not affect the distribution.

The description of $P$ allows up to $N(N-1)/2$ parameters
that affect the outcome.
Much more parsimonious descriptions can be made by
taking $P$ to be the Kronecker product of two or more
smaller matrices.
Recall that the Kronecker product of matrices $X\in\r^{m\times n}$
and $Y\in\r^{r\times s}$ is
$$
X\otimes Y \equiv 
\begin{pmatrix}
X_{11}Y & X_{12}Y & \cdots & X_{1n}Y\\
X_{21}Y & X_{22}Y & \cdots & X_{2n}Y\\
\vdots & \vdots & \ddots & \vdots\\
X_{m1}Y & X_{m2}Y & \cdots & X_{mn}Y\\
\end{pmatrix} \in\r^{mr\times ns}.
$$

An extremely parsimonious 
stochastic Kronecker graph takes
$P$ to be the $r$-fold Kronecker product 
of $\Theta = \begin{pmatrix} a & b \\ b & c\end{pmatrix},$
for $a,b,c\in[0,1]$.
That is
$$ P = P^{(r)} = \Theta \otimes \Theta \otimes \cdots\otimes\Theta \equiv \Theta^{[r]}.$$
If the power $r$ is known, then only three numbers need
to be specified, and with them
we can then simulate other graphs that are like the original.
Perhaps surprisingly, stochastic Kronecker graphs imitate
many, but of course not all, of the important
features seen in large
real world graphs.  See for example~\cite{lesk:falo:2007}.

We would like to pick parameters $a,b,c\in[0,1]$
to match the properties seen in a real and large graph.
Parameter matrices
$\Theta = \begin{pmatrix} a & b \\ b & c\end{pmatrix}$
and
$\Theta^* = \begin{pmatrix} c & b \\ b & a\end{pmatrix}$
give rise to the same graph distribution.
To force identifiability, we may assume that $a\ge c$.

\section{Moment formulas}\label{sec:moment}

The Kronecker structure in $P$ makes certain aspects of
$G$ very tractable.
For example, the number $E$ of edges in $G$ 
can be shown to have expectation
\begin{align}\label{eq:eedges}
\e(E) = \frac12\bigl((a+2b+c)^r-(a+c)^r\bigr).
\end{align}
Simply counting the edges in $G$ gives us valuable
information on the parameter vector $(a,b,c)$.
Because $E$ is a sum of independent Bernoulli random variables
we find that 
$\v(E) \le \e(E)$ and so the relative 
uncertainty $\sqrt{\v(E)}/\e(E)\le \e(E)^{-1/2}$
will be small in a graph with a large number of expected edges.

This section derives equation~\eqref{eq:eedges}
and similar formulas for the expected
number of features of various types.
The expected feature counts require
sums over various sets of nodes. Section~\ref{sec:sum}
records some summation formulas that simplify that task.
Then Section~\ref{sec:efeat} turns expected feature
counts into sums
and Section~\ref{sec:kronsimple} shows how those sums
simplify for stochastic Kronecker matrices.

\subsection{Summation formulas}\label{sec:sum}

Let $i,j,k,l\in\n$ for a finite index set
$\n$.
A plain summation sign $\sum$ represents sums
over all combinations of levels
of all the subscripting indices used.
The symbol $\sum\di$ 
includes all levels
of all indices, except for any combinations
where two or more of those indices take the same value.
In several places we find that sums are easier to
do over all levels of all indices, while the desired
sums are over unique levels.  Here we record some
formulas to translate the second type into the first.

It is elementary that
\begin{align}\label{eq:fold2}
 \sum_{ij}\di f_{ij} = \sum_{ij} f_{ij} - \sum_i f_{ii},
\end{align}
and similarly
\begin{align}\label{eq:fold3}
 \sum_{ijk}\di f_{ijk} = 
\sum_{ijk} f_{ijk} - \sum_{ij}\left( f_{ijj}+f_{iji}+f_{iij}\right)
+2\sum_if_{iii}.
\end{align}
When there are four indices, we get
\begin{equation}\label{eq:fold4}
\begin{split}
 &\sum_{ijkl}\di f_{ijkl}  = 
\sum_{ijkl} f_{ijkl} 
-\sum_{ijk}\Bigl( f_{ijki} + f_{ijkj} + f_{ijkk} + f_{ijik} + f_{ijjk} + f_{iijk}\Bigr)\\
& + \sum_{ij}
\Bigl(2\left( f_{ijjj}+f_{ijii}+f_{iiji}+f_{iiij}\right)
+f_{ijij} + f_{ijji} + f_{iijj}\Bigr)
-6\sum_i f_{iiii}.
\end{split}
\end{equation}
Equation~\eqref{eq:fold4} is more complicated than the others.
It can be proved by defining
$g_{ijk} = \sum_l f_{ijkl}-f_{ijki}-f_{ijkj}-f_{ijkk}$,
writing
$\sum\di_{ijkl}\, f_{ijkl} = \sum\di_{ijk}
g_{ijk}$  
and then applying~\eqref{eq:fold3}.

In some of our formulas below, the first index is singled out 
but the others are exchangeable.
By this we mean that
$f_{ijk} = f_{ikj}$, when there are three indices,
while
$f_{ijkl} = f_{ijlk} = f_{ikjl} = f_{iklj} = f_{iljk} = f_{ilkj}$
is the version for four indices.

When indices after the first are exchangeable, then
equation~\eqref{eq:fold3} simplifies to
\begin{align}\label{eq:fold3exch}
\sum_{ijk} f_{ijk} - \sum_{ij}\Bigl( f_{ijj}+2f_{iij}\Bigr)
+2\sum_if_{iii},
\end{align}
and equation~\eqref{eq:fold4} simplifies to
\begin{align}\label{eq:fold4exch}
\sum_{ijkl} f_{ijkl}
-3\sum_{ijk}\Bigl( f_{iijk} + f_{ijjk}\Bigr)
 + \sum_{ij}
\Bigl( 2f_{ijjj}+5f_{iijj}+4f_{iiij}\Bigr)
-6\sum_i f_{iiii}.
\end{align}

When all indices $ijk$ are exchangeable,
so that $f_{ijk} = f_{ikj} = f_{jik} = f_{jki} = f_{kij} = f_{kji}$,
then equation~\eqref{eq:fold3exch} simplifies to
\begin{align}\label{eq:fold3exch2}
\sum_{ijk} f_{ijk} - 3\sum_{ij}f_{iij}+2\sum_if_{iii}.
\end{align}

\subsection{Expected feature counts for independent edges}\label{sec:efeat}

\begin{figure}
\includegraphics[width=\hsize]{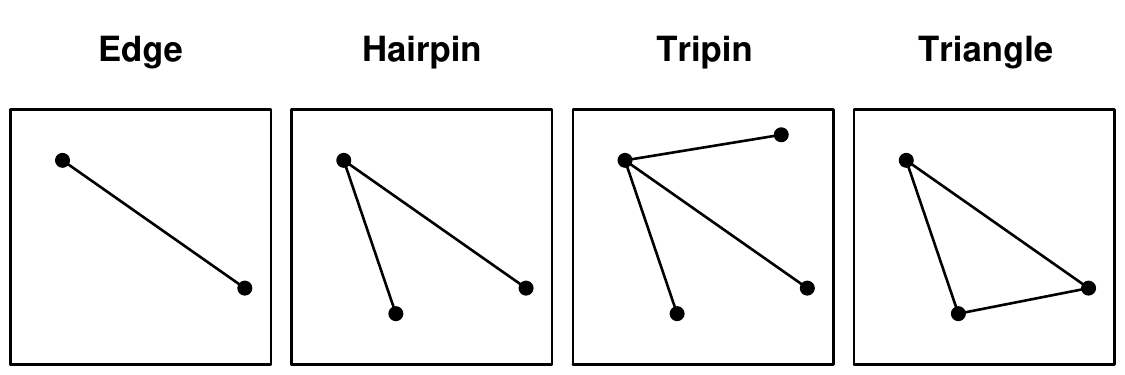}
\caption{\label{fig:feat}
This figure illustrates some of the graph features that
we can count, for use in moment based estimates
of the parameters in the stochastic Kronecker graph.
}
\end{figure}

The graph features we describe are shown
in Figure~\ref{fig:feat}. In addition
to edges, there are hairpins ($2$-stars) where two
edges share a common node, tripins ($3$-stars) where
three edges share a node, and triangles.
The Kronecker model has independent edges. Here we
find the expected feature counts for any random
graph where edge $[ij]$ appears with probability $P_{ij}$
and edges are independent.

Recall that $G^*$ is a random graph 
with $\Pr( [ij] \in G^* ) = P_{ij}$ (independently).
Let it have incidence matrix $A^*$.
There may be loops $A_{ii}^*=1$,  and for $i\ne j$,
$A_{ij}^*$ and $A_{ji}^*$ are independently generated.
The graph $G$ is formed by deleting loops from $G^*$
and symmetrizing the incidence matrix via 
$$A_{ij} = \begin{cases} A^*_{ij} & i>j\\ 0 & i=j\\ A^*_{ji} & i<j.\end{cases}$$

The number of edges in $G$
is $E = (1/2)\sum\di_{ij}A_{ij}$. The expected number of edges
satisfies
\begin{align}\label{eq:eedge}
2\, \e(E) &= \e\Bigl(\sum\di_{ij} A_{ij}\Bigr)= \sum_{ij}\di P_{ij}
= \sum_{ij}P_{ij} - \sum_iP_{ii},
\end{align}
using $\e(A_{ij})=\e(A_{ij}^*)$.

The number of hairpins in $G$ is
$H = (1/2)\sum\di_{ijk}A_{ij}A_{ik}$.
Dividing by two adjusts the sum for
counting $\{[ij],[ik]\}$ twice.
The expected value of $H$ satisfies
\begin{align*}
2\,\e(H) = & \sum_{ijk}\di P_{ij}P_{ik} = \sum_{ijk} P_{ij} P_{ik} 
-\sum_{ij}P_{ij}^2
-2\sum_{ij}P_{ii}P_{ij} +2\sum_iP_{ii}^2
\end{align*}
by letting $f_{ijk}=P_{ij}P_{ik}$,
for which $f_{ijk}=f_{ikj}$, and
applying equation~\eqref{eq:fold3exch}.

The number of triangles in $G$ is
$\Delta = (1/6)\sum\di_{ijk}A_{ij}A_{ik}A_{jk}$,
because the sum counts each triangle $3!=6$ times.
The expected value of each term is $f_{ijk} = P_{ij}P_{ik}P_{jk}$
which is symmetric in its three arguments and so
we may apply equation~\eqref{eq:fold3exch2} to get
\begin{align*}
6\,\e(\Delta) =\sum_{ijk}P_{ij}P_{ik}P_{jk}
-3\sum_{ij}P_{ii}P_{ij}^2 + 2\sum_i P_{ii}^3.
\end{align*}

The number of tripins in $G$ is
$T = (1/6)\sum\di_{ijkl}A_{ij}A_{ik}A_{il}$.
The final three indices in $f_{ijkl} = P_{ij}P_{ik}P_{il}$
are exchangeable, and so equation~\eqref{eq:fold4exch}
applies.
Thus
\begin{align*}
6\,\e(T)&=
\sum_{ijkl}P_{ij}P_{ik}P_{il}
-3\sum_{ijk}P_{ii}P_{ij}P_{ik} -3\sum_{ijk}P_{ij}^2P_{ik}\\
&+2\sum_{ij}P_{ij}^3 
+5\sum_{ij}P_{ii}P_{ij}^2
+4\sum_{ij}P_{ii}^2P_{ij}
 - 6\sum_iP_{ii}^3.
\end{align*}

\subsection{Simplifying the sums}\label{sec:kronsimple}

The sums in the expected counts simplify, because
of the properties of the Kronecker graph.
Let the node set be
$\n = \n_r = \{0,1,\dots,2^r-1\}$.
For $i\in\n$ write $i = \sum_{s=1}^{r}2^{s-1}i_s$
for $i_s\in\{0,1\}$.  Similarly let $j$, $k$, and $l$
be described in terms of $j_s,k_s,l_s\in\{0,1\}$
for $s=1,\dots,r$.

The matrix entry $P_{ij} = P^{(r)}_{ij}$ may be written
$$
P^{(r)}_{ij} = \prod_{s=1}^{r} \Theta_{i_sj_s}.
$$
For $r\ge 2$, we simplify the expression by induction
using a smaller version of the problem defined via $P^{(r-1)}$.
Specifically,
\begin{align*}
\sum_{ijk}& P^{(r)}_{ij}P^{(r)}_{ik}
 = 
\sum_{i_1}\cdots\sum_{i_{r}}
\sum_{j_1}\cdots\sum_{j_{r}}
\sum_{k_1}\cdots\sum_{k_{r}}
\prod_{s=1}^{r} \Theta_{i_sj_s}\Theta_{i_sk_s}\\
& = 
\biggl(\sum_{i_1}\cdots\sum_{i_{r-1}}
\sum_{j_1}\cdots\sum_{j_{r-1}}
\sum_{k_1}\cdots\sum_{k_{r-1}}
\prod_{s=1}^{r-1} \Theta_{i_sj_s}\Theta_{i_sk_s}\biggr)
\sum_{i_{r}j_{r}k_{r}}\Theta_{i_rj_r}\Theta_{i_rk_r}
\\
& = 
\biggl(\sum_{ijk} P^{(r-1)}_{ij}P^{(r-1)}_{ik}\biggr)
\sum_{i_{r}j_{r}k_{r}}\Theta_{i_rj_r}\Theta_{i_rk_r}\\
& = 
\biggl( \sum_{i_{r}j_{r}k_{r}}\Theta_{i_rj_r}\Theta_{i_rk_r} \biggr)^r,
\end{align*}
where indices $i_s$, $j_s$ and $k_s$ are summed over
their full ranges, and the indices $i$, $j$, $k$
for $ P^{(r-1)}_{ij}P^{(r-1)}_{ik}$ are summed over
the node set $\n_{r-1} = \{0,\dots,2^{r-1}-1\}$.

All of the sums of products of elements of $P^{(r)}_{ij}$
listed in the previous section, with summation
over all levels of each index,
also reduce this
way to $r$'th powers of their value for the
case $r=1$.

For $r=1$ we need to sum products of elements
of $P$ over $i$ or over $i,j$ or over $i,j,k$.
These cases correspond to the
first $2$, $4$, or $8$ rows of
Table~\ref{tab:ijklayout} for
$\Theta = \begin{pmatrix} a & b \\ c & d \end{pmatrix}$.
For instance $\e(H)$ requires
$\sum_{ijk}P_{ij}P_{ik}$ which we know to
be the $r$'th power of
\begin{align}
\sum_{i_{r}j_{r}k_{r}}\Theta_{i_rj_r}\Theta_{i_rk_r}
& = a^2+ba+b^2+cb+ab+b^2+bc+c^2\label{eq:thefirst}\\
& = (a+b)^2+(b+c)^2.\notag
\end{align}
The first expression~\eqref{eq:thefirst}
follows by summing over the $8$ rows of Table~\ref{tab:ijklayout}.
As a result
$$\sum_{ijk}P_{ij}P_{ik} = \bigl( (a+b)^2 + (b+c)^2\bigr)^r.$$

\begin{table}[t]
\centering
\begin{tabular}{ccc|cccc}
$i$ & $j$ & $k$ & $\Theta_{ii}$ & $\Theta_{ij}$& $\Theta_{ik}$& $\Theta_{jk}$ \\
\hline
$0$ & $0$ & $0$ & a & a & a & a\\
$1$ & $0$ & $0$ & c & b & b & a\\
$0$ & $1$ & $0$ & a & b & a & b\\
$1$ & $1$ & $0$ & c & c & b & b\\
$0$ & $0$ & $1$ & a & a & b & b\\
$1$ & $0$ & $1$ & c & b & c & b\\
$0$ & $1$ & $1$ & a & b & b & c\\
$1$ & $1$ & $1$ & c & c & c & c\\
\hline
\end{tabular}
\caption{
\label{tab:ijklayout}
This table shows entries in the matrix $\Theta$
with various indexing patterns needed in the examples.
Sums over $i$, $ij$, and $ijk$ use, respectively,
the first $2$, $4$, and $8$ rows of the table.
}
\end{table}

In the rest of this section,  we record the other sums we need.
First, the sums over one index variable take the form
\begin{align}\label{eq:sumpiipow}
\sum_i P_{ii}^m    = \bigl( a^m+c^m    \bigr)^r,
\end{align}
where cases $m=1, 2, 3$ are used in our
expected feature counts.
The sums over two index variables are
\begin{align*}
\sum_{ij} P_{ii}^mP_{ij}^n  & = 
\bigl( a^m(a^n+b^n)+c^m(b^n+c^n)\bigr)^r.
\end{align*}
The
cases we need are for $(m,n)\in\{ (0,1),(0,2),(0,3),(1,1),(1,2)\}$.

Four sums over three indices are used. They are:
\begin{align*}
\sum_{ijk} P_{ij}P_{ik} & = 
\bigl((a+b)^2 + (b+c)^2\bigr)^r\\
\sum_{ijk} P_{ij}^2P_{ik} & = 
\bigl(a^3+c^3+b(a^2+c^2)+b^2(a+c)+2b^3\bigr)^r\\
\sum_{ijk} P_{ij}P_{ik}P_{jk} & = 
\bigl( a^3+c^3 + 3b^2(a+c)\bigr)^r,\quad\text{and}\\
\sum_{ijk} P_{ii}P_{ij}P_{ik} & = 
\bigl(a(a+b)^2 + c(b+c)^2 \bigr)^r.
\end{align*}
Finally, one sum over four indices is used:
\begin{align*}
\sum_{ijkl} P_{ij}P_{ik}P_{il} & = \bigl( (a+b)^3+(b+c)^3\bigr)^r.
\end{align*}

\subsection{Expected feature counts}\label{sec:formulas}

Now we can specialize the results of Section~\ref{sec:efeat}
to the Kronecker graph setting.
Gathering together the previous developments, we find
\begin{align*}
2\,\e(E)  =\ & \bigl( a+2b+c \bigr)^r  - \bigl( a+c \bigr)^r \\
2\,\e(H)  =\ & \bigl( (a+b)^2 + (b+c)^2 \bigr)^r 
  -2\bigl( a(a+b)+ c(c+b)  \bigr)^r \\
&  -\bigl( a^2 + 2b^2 + c^2 \bigr)^r  
+ 2\bigl( a^2 + c^2 \bigr)^r \\
6\,\e(\Delta)=\ &
\bigl( a^3  + 3b^2 (a+c) + c^3  \bigr)^r  
 - 3\bigl( a(a^2+b^2) + c (b^2+c^2) \bigr)^r 
+ 2\bigl(a^3 + c^3\bigr)^r \\
6\,\e(T)=\ &
    \bigl( (a+b)^3  + (b+c)^3 \bigr)^r 
  -3\bigl( a(a+b)^2 +c(b+c)^2 \bigr)^r \\
&  -3\bigl( a^3+c^3 + b(a^2+c^2) + b^2(a+c) + 2b^3 \bigr)^r 
   +2\bigl( a^3  + 2b^3 + c^3  \bigr)^r \\
&  +5\bigl(a^3+c^3+b^2(a+c)\bigr)^r
   +4\bigl(a^3+c^3+b(a^2+c^2)\bigr)^r
   -6\bigl(a^3+c^3\bigr)^r.
\end{align*}

In each formula, the terms from sums over fewer indices
come after the ones from more indices.  The later
terms adjust for loops and double edges and other
degenerate quantities.  
For large $r$, we expect that the first term
should be most important. 
In particular if $\min(a,b,c)>0$
then in all cases the first quantity
raised to the power $r$ is the largest one.
For example the first term in $\e(E)$
is $(1+2b/(a+c))^{r}$ times as large as the second one,
which subtracts out loops. 

The first term will dominate for large $r$
unless $b\ll a+c$.
The relative magnitude of the second term is
$$\left(\frac{a+c}{a+2b+c}\right)^r
=2^{r\log_2((a+c)/(a+2b+c))}
=N^{-\alpha}
$$
where $\alpha = \log_2((a+2b+c)/(a+c))$.
If $\alpha>1/2$ then dropping the second
term in $\e(E)$ makes a smaller difference than
the sampling uncertainty in $E$.
This holds when the off diagonal element of $\Theta$
is not too small compared to the average of the diagonal elements:
$b>(\sqrt{2}-1)(a+c)/2$.

\subsection{Illustrations}

Some special cases of the formulas are
of interest.
For example if $b=0$ then there are no
edges in $G^*$ apart from loops.  As a result $G$
has $2^r$ isolated nodes.  
We find from the above
that $\e(E)=\e(H)=\e(\Delta)=\e(T)=0$ when $b=0$.

If instead $a=c=0$ then each node $i\in\n$ 
with coordinates $i_1,\dots,i_r$
has a dual node $i^*$ which
has coordinates $i^*_s=1-i_s$ for $s=1,\dots,r$.
The only possible edges in $G$ are between
nodes and their duals. There are $N=2^r$ nodes
each with probability $b^r$ of having an edge out to
its dual. The formula above gives $\e(E)=(2b)^r/2 = Nb^r/2$
when $a=c=0$, as it should.  We also get
$\e(H) = \e(\Delta)=\e(T)=0$ when $a=c=0$.

If $a=b=c=1$, then $G^*$ has every possible 
edge and loop with probability $1$.
As a result $G$ is the complete graph on $N=2^r$
nodes.  Then it has
$N(N-1)/2$ edges, $N(N-1)(N-2)/2$ hairpins, 
$N(N-1)(N-2)/2$  triangles,
and it has $N(N-1)(N-2)(N-3)/6$ tripins.

%
%
%

\section{Solving for $a$,  $b$,  and $c$}\label{sec:solving}

There are four equations in Section~\ref{sec:formulas}.
To estimate $a$, $b$, and $c$ will require at
least three of them.
Because they are high order polynomials it
is possible that there are multiple solutions
or even none at all.
The latter circumstance would provide some evidence of lack of
fit of the stochastic Kronecker model to a
given graph.  Regardless, each of the equations involves the 
count of a feature in the graph.

\subsection{Counting features in a graph}

Three of the features we use are easily obtainable
from the degrees of the nodes.
Let $d_i = \sum_{j\in\n}A_{ij}$ be the degree of node $i$
in graph $G$.
Then
\begin{align*}
E & = \frac12\sum_i d_i,\\
H & = \frac12\sum_i d_i(d_i-1),\quad\text{and}\\
T & = \frac16\sum_i d_i(d_i-1)(d_i-2)
\end{align*}
give the number of edges, hairpins (or wedges), and tripins in terms
of the degrees $d_i$.

The number of triangles $\Delta$ is not a simple function of
$d_i$.
Algorithms to count triangles are considered
in \cite{scha:wagn:2005}.
The time complexity can be as low as $O(E^{3/2})$, and
sometimes even lower for approximate 
counting~\cite{Kolountzakis-2010-triangle-counting}.

\subsection{Objective functions}

A pragmatic way to choose $a$, $b$, and $c$
is to solve
\begin{align}\label{eq:momentcrit}
\min_{a,b,c}
\sum_F \frac{(F-\e_{a,b,c}(F))^2}{\e_{a,b,c}(F)}
\end{align}
where the sum is over three or four of the
features $F\in\{E,H,T,\Delta\}$ from
Section~\ref{sec:formulas}
and the minimization is taken
over $0\le c\le a\le 1$ and $0\le b\le 1$.
The terms in~\eqref{eq:momentcrit} are scaled
by an approximate variance.
A sharper expression would account for correlations
among the features used.  That should increase
statistical efficiency, but in large problems 
lack of fit to the Kronecker model is likely
to be more important than inefficiency of estimates
within it. 

Many real world networks may not have good fits in terms of
these three Kronecker parameters.  This is the case
for most of the forthcoming experiments.  The following
more general objective can be more robust to these deviances:
\begin{align}\label{eq:generalfit}
\min_{a,b,c} \sum_F \frac{D(F,\e_{a,b,c}(F))}{N(F,\e_{a,b,c}(F))}.
\end{align}
Here $D$ is either of the two distance functions: 
\[ D_{\text{sq}}(x,y) = (x-y)^2 \quad \text{ or } \quad D_{\text{abs}}(x,y) = |x-y| \]
and $N$ is one of the normalizations: 
\[ N_{F}(F,\e) = F, \quad N_{F^2}(F,\e) = F^2, \quad 
N_{\e}(F,\e) = \e, \quad N_{\e^2}(F,\e) = \e^2. \] 
Using $D_{\text{sq}}$ and $N_{\e}$ makes it equal to the previous 
objective~\eqref{eq:momentcrit}.

In principle, either of the two distance functions
can be combined with any of the four normalizations.
We do not think it is reasonable
to expect a quadratic denominator to be a suitable
match for the absolute error.
Therefore our investigations exclude
combination of $D_{\text{abs}}$ with
either $N_{F^2}$ or $N_{\e^2}$.

We will find in Section~\ref{sec:examples} below
that robust results arise from the combination
$D_{\text{sq}}$ and $N_{F^2}$, for which~\eqref{eq:generalfit}
reduces to
\begin{align}\label{eq:ssrelerr}
\min_{a,b,c} \sum_F \Bigl(\frac{F-\e_{a,b,c}(F)}{F}\Bigr)^2,
\end{align}
a sum of squared relative errors.

Because there are only three parameters,
the criterion~\eqref{eq:generalfit} can simply
be evaluated over a grid inside
$\{ (a,b,c)\in [0,1]^3\mid a\ge c\}$.
To be sure of taking a point
within $\varepsilon$ of the minimizer
takes work $O(\varepsilon^{-3})$.
An alternative is to employ a general nonlinear
minimization procedure.
The remainder of this section looks at a method to reduce
that effort.

\subsection{Matching leading terms}
\label{sec:leading}

In a synthetic graph $N=2^r$ is  known.
When fitting to a real world graph
a pragmatic choice is $r=\lceil\log_2(N)\rceil$. 
The interpretation
is that the random graph $G^*$ may have had isolated
nodes that were then dropped when forming $G$,
but we suppose that fewer than half of the nodes in
$G^*$ have been dropped.

If we consider just the lead terms, then we 
could get estimates $\hat a$, $\hat b$, and $\hat c$
by solving three of the equations:
\begin{align*}
e\equiv (2E)^{1/r} & =\hat a+2\hat b+\hat c,\\
h \equiv(2H)^{1/r} & = (\hat a+\hat b)^2+(\hat b+\hat c)^2,\\
\delta\equiv (6\Delta)^{1/r} & = (\hat a^3+\hat c^3) +
3\hat b^2(\hat a+\hat c)\quad\text{and}\\
t \equiv (6T)^{1/r} & = (\hat a+\hat b)^3 + (\hat b+\hat c)^3.
\end{align*}

The equations for $e$ and $h$ together can be solved to get
\begin{equation}\label{eq:quad}
\begin{split}
\hat x \equiv \hat a+\hat b & = \frac{e + \sqrt{ {2 h} - {e}^{2}  }}{2}\\
\hat y \equiv \hat b+\hat c & = \frac{e - \sqrt{ {2 h} - {e}^{2}  }}{2},
\end{split}
\end{equation}
where we have assumed that $a\ge c$.
The transformed tripin count $t$ matches $\hat x^3+\hat y^3$
and so it is redundant given $e$ and $h$, if we are just using 
lead terms. We must either count triangles, or use higher order
terms.

Equation~\eqref{eq:quad} may fail to have a meaningful
solution.  At a minimum we require
$e^2 \le 2h$ and $e\ge\sqrt{2h-e^2}$. These translate into
$$h\le e^2 \le 2h,$$
which is equivalent to
$$2H\le 4E^2 \le 2^{r+1}H$$
that in terms of node degrees is
\begin{align}\label{eq:okbydeg}
\sum_i d_i(d_i-1) \le \Bigl(\sum_id_i\Bigr)^2 \le N\sum_id_i(d_i-1).
\end{align}
The left hand inequality in~\eqref{eq:okbydeg} holds for
any graph, but the right hand side need not. It holds
when
$N^{-1}\sum_i(d_i-\bar d)^2\ge \bar d = N^{-1}\sum_id_i$.
If the variance of the node degrees $d_i$ is smaller than their
mean, then equation~\eqref{eq:quad} does not have real
valued solutions.
The degree distribution of a stochastic Kronecker
graph has heavy tails \cite{mahd:xu:2007}.
Therefore in applications where that model is suitable
equation~\eqref{eq:quad} will
give a reasonable solution.

When $d_i$ have a variance larger than their
mean, then we can do a univariate grid search
for $b\in[0,1]$
using equation~\eqref{eq:quad} to get
$a = x-b\equiv a(b)$ and $c=y-b\equiv c(b)$.
The choice of $b$ can then be
made as the minimizer of
$|a(b)^3+c(b)^3+3b^2(a(b)+c(b))-\delta|$.

\section{Examples}\label{sec:examples}
In this section, we experiment with different techniques for
fitting the parameters of the Kronecker model.  These experiments
involve 8 real world networks whose statistical properties are
listed in the rows of the forthcoming tables labeled ``Source.''

The networks ca-GrQc, ca-HepTh, ca-HepPh are co-authorship networks 
from arXiv~\cite{Leskovec-2007-densification}.  The nodes of the network represent 
authors, and there is an edge between two nodes when the authors jointly wrote a paper. 
Likewise, the hollywood-2009 network is a collaboration graph between actors and 
actresses in IMDB~\cite{Boldi-2011-layered,boldi2004-webgraph}.  Nodes are actresses or 
actors, and edges are collaborations on a movie, as evidenced by jointly appearing on 
the cast.  These networks are naturally undirected and all edges are unweighted.  

Both as20000102 and as-Skitter are technological infrastructure 
networks~\cite{Leskovec-2007-densification}.  Each node represents a 
router on the internet and edges represent 
a physical or virtual connection between the routers.  Again, these networks are 
undirected and unweighted.  

The wikipedia-20051105 graph is a symmetrized link graph of 
the articles on Wikipedia generated from a data download on November 5th, 2005~\cite{%
constantine2007-pagerank-pce}.  The underlying network is directed, but in these 
experiments, we have converted it into an undirected network by dropping the direction 
of the edges.  

All of the previously described networks have distinctly skewed degree distributions.  
That is, there are a few authors, actors, routers, or articles with a large number of 
links, despite the overall network having a small average degree.  The final network we 
study is usroads, a highway-level network from the National Highway Planning Network 
(\url{http://www.fhwa.dot.gov/planning/nhpn/}), which does not have a highly skewed 
distribution.  We include it as an example of a nearly planar network.  It is also
naturally undirected.

In two of the experiments, we generate synthetic Kronecker networks.  
The algorithm to realize these networks is an explicit coin-flipping
procedure instead of the more common ball-dropping method 
\cite{Leskovec-2010-KronFit}.  For each cell $i,j$ in the $2^r-1 \choose 2$
upper triangular portion, we first determine the 
log of the probability of a non-zero value in that cell, then generate 
a random coin flip with that probability as heads and record an edge when the coin 
comes up heads.  This procedure is scalable because the full matrix of probabilities is 
never formed.  It is also easily 
parallelizable.  Our implementation uses pthreads to exploit multi-core
parallelism.  It takes somewhat more work than the ball-dropping
procedure, scaling as $O(r2^{2r})$ instead of $O(rm)$, where $m$ 
is the number of balls dropped. Often $m \approx 2^{r+3}$, that is, 
$8$ balls per vertex~\cite{Groer-2011-rmat}.  Each ball generates
about one edge; see \cite{Groer-2011-rmat} for a more thorough analysis.  
Coin-flipping preserves the exact Kronecker distribution whereas
ball-dropping is an approximation.

The experiments with these networks investigate (i) the difference in results from
the various choices of $D$ and $N$ in the objective~\eqref{eq:generalfit}; (ii)
the fitted parameters to the 8 real world networks;
and (iii) the difference
in fitted parameters when only using three of the four graph features.

\subsection{Objective functions}

The first study regards the choice of objective function.  
Of eight possible combinations of distance and
normalization, we considered two to be unreasonable
a priori.  Here we investigate the other six pairs.

Table~\ref{tab:objective} shows the different parameters
$a,b,$ and $c$ chosen by each objective function, as well as the expected
feature counts for those parameters for three graphs:
a single realization of a Kronecker graph with $a=0.99,b=0.48,c=0.25$,
the collaboration network ca-GrQc, and the infractucture network as20000102. 
The rows labeled
``Source'' contain the actual feature counts in each network.
The optimization algorithm to pick $a,b,c$ uses the best objective value from three 
procedures.  First, it tries 50 random starting points for the \text{fmincon} function 
in Matlab R2010b, an active set algorithm.  Then, it performs a grid search procedure 
with 100 equally spaced points in each dimension.  Finally, it tries
the leading term matching algorithm from Section~\ref{sec:leading}, and
considers those parameters.

The results in the table show that the choice of objective function does not make a 
difference when 
the graph fits the Kronecker model.  However, it can
make a large difference when the graph does not exactly fit, as in the
ca-GrQc and as20000102 networks.  Both of the objectives $D_{\text{sq}},N_{\e^2}$ and 
$D_{\text{abs}},N_{\e}$ produced distinctly different fits for these two networks, compared to the other objectives.  These two 
fits seem to be primarily matching the number of triangles -- almost to the exclusion 
of the other features.  The other odd fit for the ca-GrQc graph comes from the 
$D_{\text{sq}},N_{\e}$ 
objective.  This fit appears to be matching the tripin count and
ignoring other features, something that also seems to be true for the as20000102 
graph.  Among the remaining fits for ca-GrQc, there is little difference among the 
fitted parameters and estimated features.  The results are a bit different for 
as20000102.  The fits for $D_{\text{sq}},N_{\e}$ and $D_{\text{sq}},N_{F}$ are almost 
identical and show a good match to the tripin count, but a poor match to the remaining 
features.  The fits for $D_{\text{sq}},N_{F^2}$ and $D_{\text{abs}},N_{F}$ are similar 
and deciding which is better seems like a matter of preference.
These observations held up under further experimentation, which we omit here in the 
interest of space.

Based on these results, either of the objectives $D_{\text{sq}},N_{F^2}$ or 
$D_{\text{abs}},N_F$ appears to be a robust choice when the model does not fit 
exactly.  Due to the continuity of the $D_{\text{sq}}$ function, the rest of our fits 
in this manuscript uses the $D_{\text{sq}},N_{F^2}$ variation.

%

\begin{table}
\caption{For three graphs, the fitted Kronecker parameters $a,b,c$ for variations on the objective function~\eqref{eq:generalfit}.  Subsequent columns show 
counts for these parameters; the row labeled Source shows the actual network feature values $F_{\mathrm{obs}}$.
The other rows show $\e(F)/F_{\mathrm{obs}}$.
The objective column shows the value of the objective function at the minimizer.}
\smallskip
\centering
\label{tab:objective}

\setlength{\tabcolsep}{2pt}
\footnotesize
\begin{tabularx}{\linewidth}{llll*{5}{>{\raggedleft}X}l}
\toprule
\textbf{Graph} & \multicolumn{3}{l}{Kron. Parameters}& \multicolumn{5}{l}{Graph /
Expected Features} & Obj. \\
Fit type & $a$ & $b$ & $c$ & Verts. & Edges & Hairpins & Tripins & Tris. & \\
\midrule
\multicolumn{4}{l}{ \bfseries Stochastic Kronecker} \\ 
   Source &    0.99 &   0.48 &   0.25 &           16384 &           30830 &          521676 &         8659050 &             854 & --- \\ 
$D_{\text{sq}}, N_{\e}$ &  0.993 &  0.476 &  0.255 &           16384 &            1.00 &            1.00 &           1.000 &          1.0010 & $7.76\tdot 10^{\mbox{-}1}$ \\ 
$D_{\text{sq}}, N_{\e^2}$ &  0.993 &  0.476 &  0.254 &      \dupcolval &            1.00 &            1.00 &           1.001 &          1.0000 & $9.72\tdot 10^{\mbox{-}6}$ \\ 
$D_{\text{sq}}, N_{F}$ &  0.993 &  0.476 &  0.255 &      \dupcolval &            1.00 &            1.00 &           1.000 &          1.0014 & $7.80\tdot 10^{\mbox{-}1}$ \\ 
$D_{\text{sq}}, N_{F^2}$ &  0.993 &  0.476 &  0.254 &      \dupcolval &            1.00 &            1.00 &           1.001 &          1.0000 & $9.71\tdot 10^{\mbox{-}6}$ \\ 
$D_{\text{abs}}, N_{\e}$ &  0.993 &  0.476 &  0.253 &      \dupcolval &            1.00 &            1.00 &           1.000 &          1.0000 & $4.19\tdot 10^{\mbox{-}3}$ \\ 
$D_{\text{abs}}, N_{F}$ &  0.993 &  0.476 &  0.253 &      \dupcolval &            1.00 &            1.00 &           1.000 &          1.0000 & $4.17\tdot 10^{\mbox{-}3}$ \\ 
  Leading &  0.990 &  0.479 &  0.250 &      \dupcolval &            1.00 &            1.00 &           1.006 &          0.9835 & --- \\ 
\midrule 
\multicolumn{2}{l}{ \bfseries              ca-GrQc} \\ 
   Source &    --- &    --- &    --- &            5242 &           14484 &          229867 &         2482738 &           48260 & --- \\ 
$D_{\text{sq}}, N_{\e}$ &  1.000 &  0.221 &  1.000 &            8192 &            3.52 &            2.74 &           1.028 &          0.0666 & $9.14\tdot 10^{5}$ \\ 
$D_{\text{sq}}, N_{\e^2}$ &  1.000 &  0.733 &  0.000 &      \dupcolval &            4.30 &           29.82 &         355.084 &          0.9052 & $2.53\tdot 10^{0}$ \\ 
$D_{\text{sq}}, N_{F}$ &  1.000 &  0.459 &  0.312 &      \dupcolval &            1.17 &            0.99 &           1.001 &          0.0107 & $4.77\tdot 10^{4}$ \\ 
$D_{\text{sq}}, N_{F^2}$ &  1.000 &  0.467 &  0.279 &      \dupcolval &            1.06 &            0.92 &           1.035 &          0.0107 & $9.89\tdot 10^{\mbox{-}1}$ \\ 
$D_{\text{abs}}, N_{\e}$ &  1.000 &  0.737 &  0.000 &      \dupcolval &            4.51 &           32.38 &         397.213 &          1.0000 & $2.75\tdot 10^{0}$ \\ 
$D_{\text{abs}}, N_{F}$ &  1.000 &  0.469 &  0.267 &      \dupcolval &            1.00 &            0.87 &           1.000 &          0.0103 & $1.12\tdot 10^{0}$ \\ 
  Leading &  1.000 &  0.488 &  0.229 &      \dupcolval &            1.00 &            1.00 &           1.405 &          0.0131 & --- \\ 
\midrule 
\multicolumn{2}{l}{ \bfseries           as20000102} \\ 
   Source &    --- &    --- &    --- &            6474 &           12572 &         2059364 & $6.75\tdot 10^{8}$ &            6584 & --- \\ 
$D_{\text{sq}}, N_{\e}$ &  1.000 &  0.722 &  0.000 &            8192 &            4.42 &            2.73 &           0.997 &          5.2222 & $2.32\tdot 10^{6}$ \\ 
$D_{\text{sq}}, N_{\e^2}$ &  0.712 &  0.947 &  0.000 &      \dupcolval &           10.13 &            4.89 &           0.840 &          1.1082 & $1.49\tdot 10^{0}$ \\ 
$D_{\text{sq}}, N_{F}$ &  1.000 &  0.722 &  0.000 &      \dupcolval &            4.40 &            2.71 &           0.989 &          5.1843 & $6.39\tdot 10^{6}$ \\ 
$D_{\text{sq}}, N_{F^2}$ &  1.000 &  0.632 &  0.000 &      \dupcolval &            1.63 &            0.51 &           0.101 &          0.7029 & $1.54\tdot 10^{0}$ \\ 
$D_{\text{abs}}, N_{\e}$ &  0.676 &  0.980 &  0.000 &      \dupcolval &           11.83 &            5.98 &           1.000 &          1.0000 & $1.75\tdot 10^{0}$ \\ 
$D_{\text{abs}}, N_{F}$ &  1.000 &  0.648 &  0.000 &      \dupcolval &            1.95 &            0.68 &           0.152 &          1.0000 & $2.12\tdot 10^{0}$ \\ 
\bottomrule 
\end{tabularx}

\end{table}

\subsection{Parameters for real-world networks}

For the 8 networks previously described, we use the objective function 
\eqref{eq:generalfit} with $D_{\text{sq}}, N_{F^2}$ to fit the parameters $a,b,c$.  The 
results, along with the expected feature counts for the fitted parameters, are 
presented in Table~\ref{tab:data-and-fits}.  We show the minimizer for the three different 
strategies to optimize the objective described in the previous section: a direct 
minimization procedure, the grid search procedure, and the leading term matching 
approach (Section~\ref{sec:leading}).  For each approach, the table also shows the time 
required for that algorithm and the value of the objective function at the minimizer.  

Leskovec et al.~\cite{Leskovec-2010-KronFit} provide the fitted parameters $a,b,$ and $c
$ from their KronFit algorithm for the networks ca-GrQc, ca-HepTh, ca-HepPh, and 
as20000102.  We include them in Table~\ref{tab:data-and-fits} for comparison.
In all cases but one, the expected feature count
using KronFit is farther from the observed feature
count than the expectation under our moment based fits.
Sometimes it is much farther.
There was one exception.
For the graph as20000102, KronFit gave a better
estimate of the number of edges than our moment
method gave.

KronFit typically underestimates the feature counts.
The effect is severe for triangles.  Kronecker
random graphs commonly have many fewer triangles than
the real world graphs to which they are fit.  
Our moment based estimators
find parameters leading to many more triangles than
the KronFit parameters do.

In fairness, we point out that our method is designed
to match expected to observed feature counts,
while KronFit fits by maximum likelihood.  Therefore
the evaluation criterion is closer to the fitting
criterion for us.
But maximum likelihood ordinarily beats or matches
the method of moments in large samples from parametric models;
it's mismatching criteria are more than compensated for
by superior statistical efficiency.
The explanation here may involve maximum likelihood being less
robust to lack of fit of the Kronecker model, or 
it may be that KronFit is not finding the MLE.

The results in Table~\ref{tab:data-and-fits} show small differences in the fits between the direct and grid algorithms, 
although the direct algorithm is much faster.  The leading term matching algorithm, 
when it succeeds, generates similar Kronecker parameters, although with a distinctly 
worse objective value. 
The results from the KronFit algorithm differ and likely match 
the graph in another aspect.  

Lead term matching is tens of times faster
than direct search and roughly $1000$ times faster
than grid search.  But even the grid search takes under a minute in our examples,
so the speed savings from the lead term approach is of little benefit here.
For the large graphs, the time to compute the network 
features dominates the time to fit the parameters, showing that this approach 
scales to large networks.  

Overall, the results indicate that the Kronecker models tend not to be a good fit to 
the data.  The model appears to have a considerable difference in at least once of the 
graph features.  Usually, it's the number of triangles, which differs by up to two 
orders of magnitude for many of the collaboration networks.  

\begin{table}
\caption{
The fitted Kronecker parameters for variations on the algorithm
-- direct, grid, leading, or 
KronFit~\cite{Leskovec-2010-KronFit} --
to minimize of the objective function~\eqref{eq:generalfit}
Subsequent columns show the expected feature 
counts for these parameters; the row labeled Source shows the actual network features.
The 
time column is either the time to compute the features on the original
graph or the time for the algorithm to fit the parameters.}
\label{tab:data-and-fits}

\setlength{\tabcolsep}{2pt}
\footnotesize
\begin{tabularx}{\linewidth}{llll*{5}{>{\raggedleft}X}ll}
\toprule
\textbf{Graph} & \multicolumn{3}{l}{Kron. Parameters}& \multicolumn{5}{l}{Graph /
Expected Features} & & \rlap{Time} \\
Fit type & $a$ & $b$ & $c$ & Verts. & Edges & Hairpins & Tripins & Tris. & Obj. & 
\rlap{(sec.)} \\
\midrule
\multicolumn{2}{l}{ \bfseries              ca-GrQc} \\ 
   Source &    --- &    --- &    --- &            5242 &           14484 &          229867 &         2482738 &           48260 & --- & $<\!\!0.05$\\ 
   Direct &  1.000 &  0.467 &  0.279 &            8192 &            1.06 &            0.92 &           1.035 &          0.0107 & 0.989 &    1.0\\ 
     Grid &  1.000 &  0.470 &  0.270 &      \dupcolval &            1.03 &            0.91 &           1.060 &          0.0108 & 0.991 &   48.5\\ 
  Leading &  1.000 &  0.488 &  0.229 &      \dupcolval &            1.00 &            1.00 &           1.405 &          0.0131 & 1.138 & $<\!\!0.05$\\ 
  KronFit &  0.999 &  0.245 &  0.691 &      \dupcolval &            0.84 &            0.20 &           0.029 &          0.0012 & 2.935 &    --- \\ 
\midrule 
\multicolumn{2}{l}{ \bfseries             ca-HepPh} \\ 
   Source &    --- &    --- &    --- &           12008 &          118489 &        15278011 & $1.28\tdot 10^{9}$ &         3358499 & --- &    1.9\\ 
   Direct &  1.000 &  0.669 &  0.101 &           16384 &            1.11 &            0.82 &           1.064 &          0.0164 & 1.015 &    0.8\\ 
     Grid &  1.000 &  0.670 &  0.100 &      \dupcolval &            1.12 &            0.84 &           1.091 &          0.0167 & 1.016 &   48.6\\ 
  Leading &  1.000 &  0.708 &  0.005 &      \dupcolval &            1.00 &            1.00 &           2.021 &          0.0196 & 2.004 & $<\!\!0.05$\\ 
  KronFit &  0.999 &  0.437 &  0.484 &      \dupcolval &            0.69 &            0.10 &           0.014 &          0.0006 & 3.196 &    --- \\ 
\midrule 
\multicolumn{2}{l}{ \bfseries             ca-HepTh} \\ 
   Source &    --- &    --- &    --- &            9877 &           25973 &          299356 &         2098335 &           28339 & --- & $<\!\!0.05$\\ 
   Direct &  1.000 &  0.401 &  0.379 &           16384 &            1.06 &            0.92 &           1.035 &          0.0112 & 0.989 &    0.8\\ 
     Grid &  1.000 &  0.400 &  0.380 &      \dupcolval &            1.05 &            0.90 &           1.001 &          0.0109 & 0.991 &   48.7\\ 
  Leading &  1.000 &  0.423 &  0.325 &      \dupcolval &            1.00 &            1.00 &           1.444 &          0.0140 & 1.169 & $<\!\!0.05$\\ 
  KronFit &  0.999 &  0.271 &  0.587 &      \dupcolval &            0.74 &            0.25 &           0.073 &          0.0020 & 2.936 &    --- \\ 
\midrule 
\multicolumn{2}{l}{ \bfseries       hollywood} \\ 
   Source &    --- &    --- &    --- &         1139905 &        56375711 & $4.76\tdot 10^{10}$ & $3.24\tdot 10^{13}$ & $4.92\tdot 10^{9}$ & --- & 2946.1\\ 
   Direct &  1.000 &  0.623 &  0.186 &         2097152 &            1.13 &            0.76 &           1.070 &          0.0029 & 1.075 &    1.0\\ 
     Grid &  1.000 &  0.620 &  0.200 &      \dupcolval &            1.21 &            0.80 &           1.055 &          0.0030 & 1.083 &   48.6\\ 
  Leading &  1.000 &  0.662 &  0.095 &      \dupcolval &            1.00 &            1.00 &           2.670 &          0.0046 & 3.779 & $<\!\!0.05$\\ 
\midrule 
\multicolumn{2}{l}{ \bfseries           as20000102} \\ 
   Source &    --- &    --- &    --- &            6474 &           12572 &         2059364 & $6.75\tdot 10^{8}$ &            6584 & --- & $<\!\!0.05$\\ 
   Direct &  1.000 &  0.632 &  0.000 &            8192 &            1.63 &            0.51 &           0.101 &          0.7029 & 1.541 &    0.8\\ 
     Grid &  1.000 &  0.630 &  0.000 &      \dupcolval &            1.60 &            0.49 &           0.096 &          0.6717 & 1.543 &   48.7\\ 
  KronFit &  0.987 &  0.571 &  0.049 &      \dupcolval &            0.99 &            0.17 &           0.018 &          0.1738 & 2.655 &    --- \\ 
\midrule 
\multicolumn{2}{l}{ \bfseries           as-skitter} \\ 
   Source &    --- &    --- &    --- &         1696415 &        11095298 & $1.60\tdot 10^{10}$ & $9.66\tdot 10^{13}$ &        28769868 & --- &  107.0\\ 
   Direct &  1.000 &  0.644 &  0.000 &         2097152 &            1.61 &            0.74 &           0.239 &          0.1384 & 1.755 &    0.7\\ 
     Grid &  1.000 &  0.640 &  0.000 &      \dupcolval &            1.48 &            0.65 &           0.199 &          0.1181 & 1.776 &   48.7\\ 
\midrule 
\multicolumn{2}{l}{ \bfseries   wiki-2005} \\ 
   Source &    --- &    --- &    --- &         1634989 &        18540603 & $3.72\tdot 10^{10}$ & $3.72\tdot 10^{14}$ &        44667105 & --- &  378.9\\ 
   Direct &  1.000 &  0.674 &  0.000 &         2097152 &            1.64 &            0.79 &           0.211 &          0.2589 & 1.629 &    0.6\\ 
     Grid &  1.000 &  0.670 &  0.000 &      \dupcolval &            1.53 &            0.70 &           0.179 &          0.2246 & 1.646 &   48.5\\ 
\midrule 
\multicolumn{2}{l}{ \bfseries           usroads} \\ 
   Source &    --- &    --- &    --- &          126146 &          161950 &          292425 &          115885 &            4113 & --- & $<\!\!0.05$\\ 
   Direct &  1.000 &  0.070 &  1.000 &          131072 &            0.88 &            1.04 &           1.057 &          0.1177 & 0.798 &    1.0\\ 
     Grid &  1.000 &  0.070 &  1.000 &      \dupcolval &            0.87 &            1.03 &           1.012 &          0.1148 & 0.800 &   48.5\\ 
\bottomrule 
\end{tabularx}
\end{table}

\subsection{Fitting partial sets of features}

The previous set of experiments illustrated that the Kronecker graphs may not 
simultaneously fit all four of the network features: edges, hairpins/wedges, tripins, and 
triangles.  In Table~\ref{tab:partial-fits}, we examine the change in fits when only 
using three of the four network features in the summation in the 
objective~\eqref{eq:generalfit}.  
We take the set of parameters with the smallest objective among all 
the procedures investigated in the previous section.  The results show small changes to 
the parameters and expected feature fits.  Nonetheless, the minimizer remains mostly 
unchanged.

\begin{table}
\caption{The change in fitted parameters when the objective 
function~\eqref{eq:generalfit} only considers three of the 
four features.   The row labeled ``-Tris'',
for instance, gives the fitted parameters when triangles are not 
included in~\eqref{eq:momentcrit}.  Rows labeled ``source'' 
again contain the actual graph features, and the rows labeled ``all''
show the parameters fitted to all four features.  }
\label{tab:partial-fits}

\setlength{\tabcolsep}{2pt}
\footnotesize
\begin{tabularx}{\linewidth}{llll*{5}{>{\raggedleft}X}ll}
\toprule
\textbf{Graph} & \multicolumn{3}{l}{Kron. Parameters}& \multicolumn{5}{l}{Graph /
Expected Features} & & \rlap{Time} \\
Fit type & $a$ & $b$ & $c$ & Verts. & Edges & Hairpins & Tripins & Tris. & Obj. & \rlap{(sec.)} \\
\midrule 
\multicolumn{2}{l}{ \bfseries              ca-GrQc} \\ 
   Source &    --- &    --- &    --- &            5242 &           14484 &          229867 &         2482738 &           48260 & --- & $<\!\!0.05$\\ 
\addlinespace 
      All &  1.000 &  0.467 &  0.279 &            8192 &            1.06 &            0.92 &           1.035 &          0.0107 & 0.989 &   54.8\\ 
  KronFit &  0.999 &  0.245 &  0.691 &      \dupcolval &            0.84 &            0.20 &           0.029 &          0.0012 & 2.935 &    --- \\ 
\addlinespace 
   -Edges &  1.000 &  0.458 &  0.317 &      \dupcolval &            1.19 &            1.00 &           1.007 &          0.0108 & 0.978 &   53.9\\ 
-Hairpins &  1.000 &  0.469 &  0.267 &      \dupcolval &            1.00 &            0.87 &           1.007 &          0.0103 & 0.980 &   53.9\\ 
 -Tripins &  1.000 &  0.493 &  0.216 &      \dupcolval &            0.99 &            1.02 &           1.536 &          0.0139 & 0.973 &   54.0\\ 
    -Tris &  1.000 &  0.467 &  0.279 &      \dupcolval &            1.06 &            0.92 &           1.029 &          0.0106 & 0.011 &   56.1\\ 
\midrule 
\multicolumn{2}{l}{ \bfseries             ca-HepPh} \\ 
   Source &    --- &    --- &    --- &           12008 &          118489 &        15278011 & $1.28\tdot 10^{9}$ &         3358499 & --- &    1.9\\ 
\addlinespace 
      All &  1.000 &  0.669 &  0.101 &           16384 &            1.11 &            0.82 &           1.064 &          0.0164 & 1.015 &   54.1\\ 
  KronFit &  0.999 &  0.437 &  0.484 &      \dupcolval &            0.69 &            0.10 &           0.014 &          0.0006 & 3.196 &    --- \\ 
\addlinespace 
   -Edges &  1.000 &  0.650 &  0.192 &      \dupcolval &            1.49 &            1.02 &           1.006 &          0.0201 & 0.960 &   57.2\\ 
-Hairpins &  1.000 &  0.670 &  0.083 &      \dupcolval &            1.01 &            0.75 &           1.007 &          0.0146 & 0.971 &   57.2\\ 
 -Tripins &  1.000 &  0.709 &  0.005 &      \dupcolval &            1.01 &            1.02 &           2.065 &          0.0200 & 0.961 &   56.9\\ 
    -Tris &  1.000 &  0.669 &  0.099 &      \dupcolval &            1.10 &            0.82 &           1.058 &          0.0162 & 0.047 &   54.7\\ 
\midrule 
\multicolumn{2}{l}{ \bfseries             ca-HepTh} \\ 
   Source &    --- &    --- &    --- &            9877 &           25973 &          299356 &         2098335 &           28339 & --- & $<\!\!0.05$\\ 
\addlinespace 
      All &  1.000 &  0.401 &  0.379 &           16384 &            1.06 &            0.92 &           1.035 &          0.0112 & 0.989 &   57.4\\ 
  KronFit &  0.999 &  0.271 &  0.587 &      \dupcolval &            0.74 &            0.25 &           0.073 &          0.0020 & 2.936 &    --- \\ 
\addlinespace 
   -Edges &  1.000 &  0.391 &  0.417 &      \dupcolval &            1.19 &            1.00 &           1.006 &          0.0114 & 0.977 &   56.5\\ 
-Hairpins &  1.000 &  0.404 &  0.365 &      \dupcolval &            1.00 &            0.87 &           1.008 &          0.0108 & 0.979 &   57.1\\ 
 -Tripins &  1.000 &  0.431 &  0.308 &      \dupcolval &            0.98 &            1.03 &           1.623 &          0.0152 & 0.971 &   56.9\\ 
    -Tris &  1.000 &  0.401 &  0.379 &      \dupcolval &            1.06 &            0.92 &           1.028 &          0.0111 & 0.011 &   56.6\\ 
\midrule 
\multicolumn{2}{l}{ \bfseries           as20000102} \\ 
   Source &    --- &    --- &    --- &            6474 &           12572 &         2059364 & $6.75\tdot 10^{8}$ &            6584 & --- & $<\!\!0.05$\\ 
\addlinespace 
      All &  1.000 &  0.632 &  0.000 &            8192 &            1.63 &            0.51 &           0.101 &          0.7029 & 1.541 &   56.7\\ 
  KronFit &  0.987 &  0.571 &  0.049 &      \dupcolval &            0.99 &            0.17 &           0.018 &          0.1738 & 2.655 &    --- \\ 
\addlinespace 
   -Edges &  0.935 &  0.720 &  0.000 &      \dupcolval &            3.04 &            1.12 &           0.235 &          1.0796 & 0.608 &   57.3\\ 
-Hairpins &  1.000 &  0.621 &  0.000 &      \dupcolval &            1.44 &            0.41 &           0.077 &          0.5526 & 1.250 &   58.5\\ 
 -Tripins &  1.000 &  0.628 &  0.000 &      \dupcolval &            1.56 &            0.47 &           0.091 &          0.6400 & 0.723 &   56.7\\ 
    -Tris &  1.000 &  0.618 &  0.000 &      \dupcolval &            1.39 &            0.39 &           0.071 &          0.5137 & 1.392 &   57.1\\ 
\bottomrule
\end{tabularx}
\end{table}

Table~\ref{tab:partial-fits} provides a kind of cross-validated
feature estimation, showing the accuracy of a feature's
estimate when it is not included in the fitting.  
Apart from the exception noted before
(the edge counts for as20000102)
our moment based estimates give closer
matches to the source feature counts
than KronFit provides, whether 
the moment being studied is part of the fitting process or not.

We see some examples where leaving out one feature
seems to improve the fitting of another. For instance,
in three of the four graphs, leaving out the tripin
count improved the match for triangles and conversely.

\section{Synthetic examples}\label{sec:simu}

The results from the previous section show that there can often by a large deviation in 
the expected moments of the best Kronecker fit.  In this section, we investigate the 
accuracy of the fitting procedure when the graph is a realization of a stochastic 
Kronecker network.  

\begin{figure}
\centering
\subfigure[Kronecker $0.99,0.48,0.25$, $r=14$]{\includegraphics[height=1.5in]{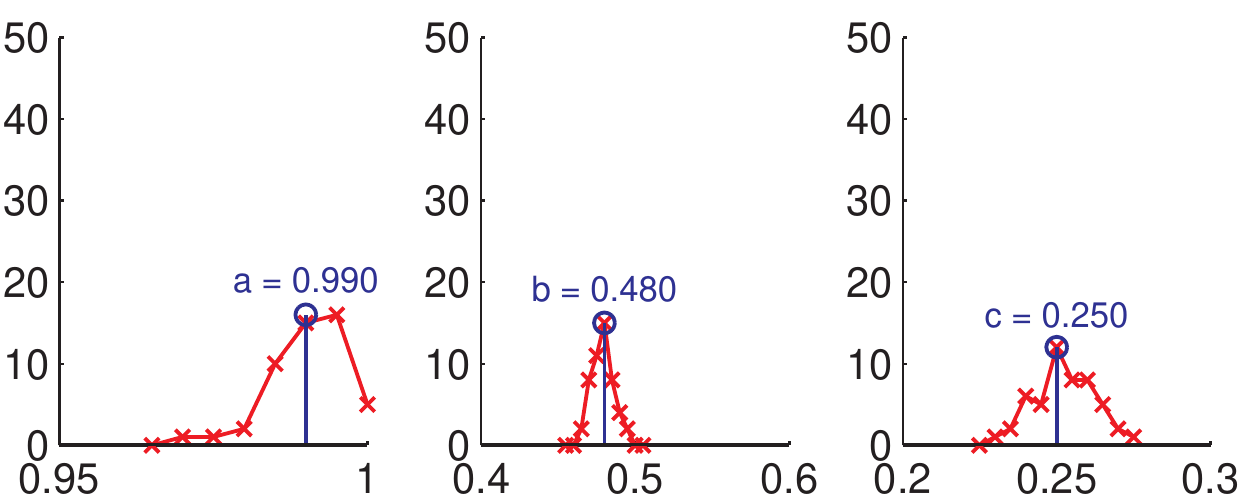}}
\subfigure[Kronecker $1.0,0.67,0.08$, $r=14$]{\includegraphics[height=1.5in]{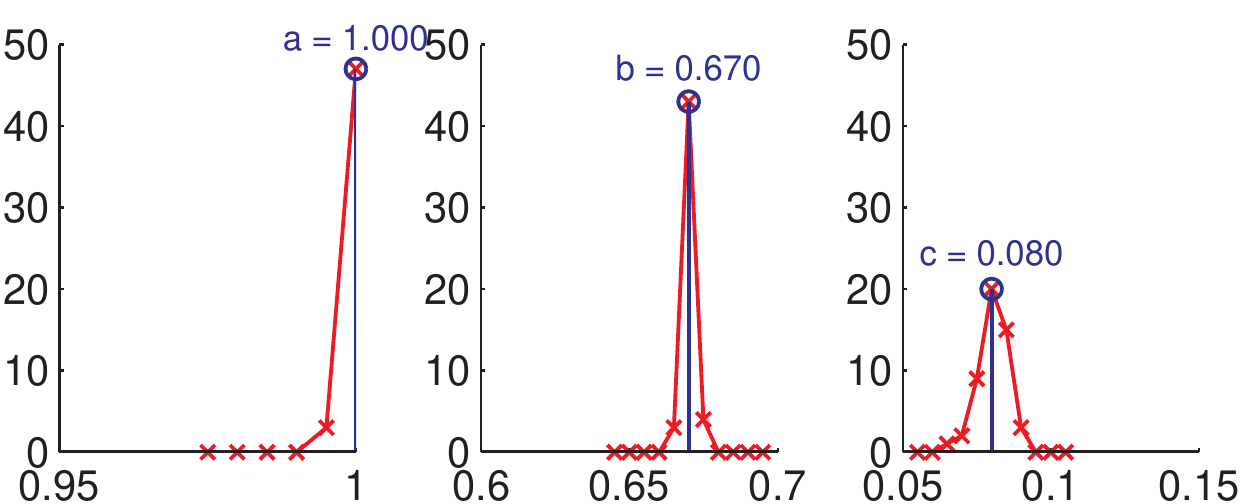}}
\subfigure[Kronecker $0.999,0.271,0.587$, $r=14$]{\includegraphics[height=1.5in]{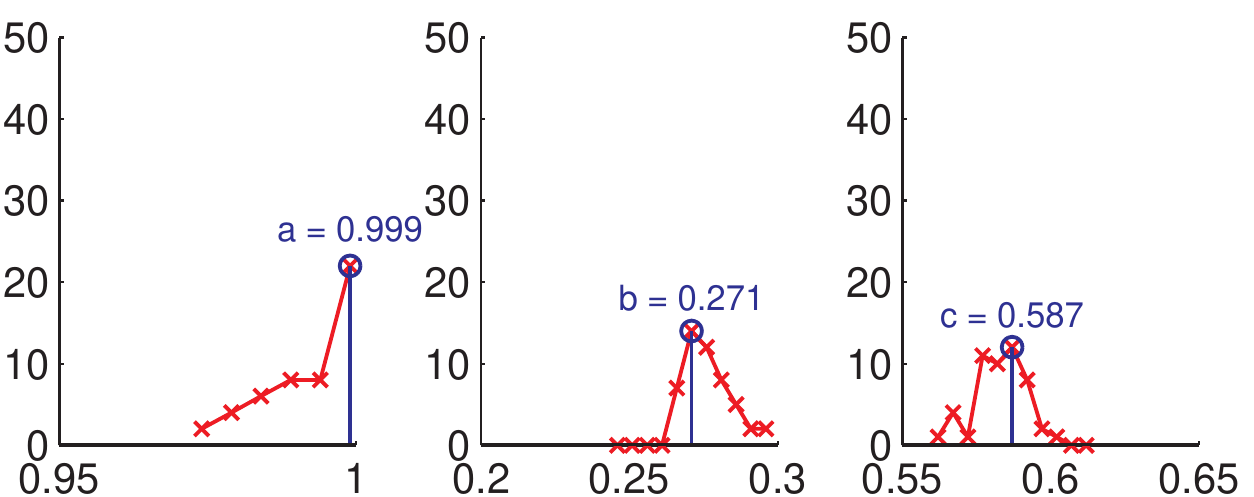}}
\subfigure[Kronecker $0.87,0.6,0.7$, $r=14$]{\includegraphics[height=1.5in]{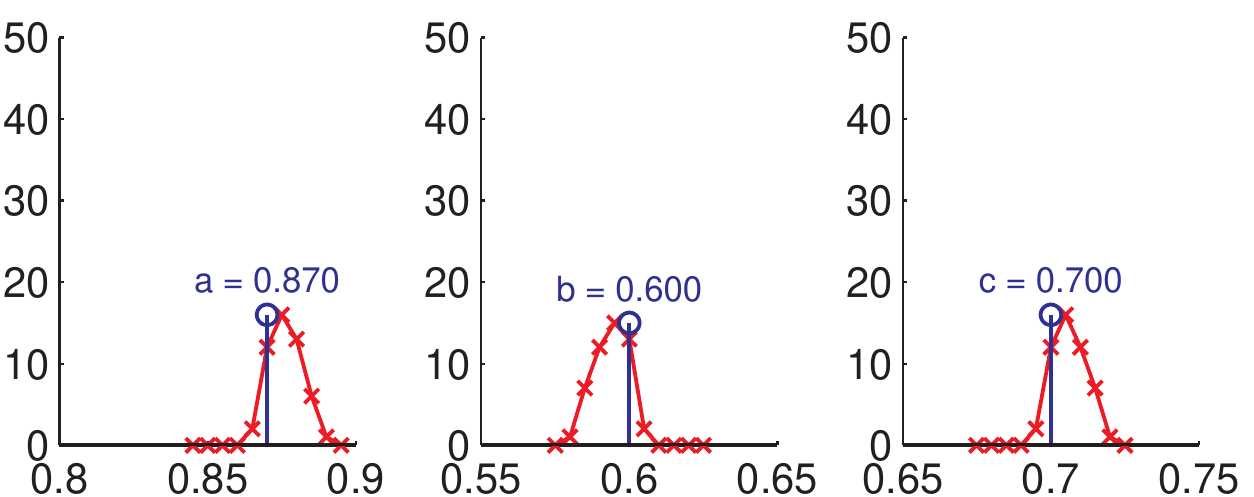}}
\caption{Histograms of fitted parameters to 50 realizations
of a Kronecker graphs with the parameters given in the caption (red lines).}
\label{fig:kron-fits}
\end{figure}

For four sets of Kronecker parameters:
\begin{compactitem}
\item $(a,b,c) = (0.99,0.48,0.25)$, $r=14$
\item $(a,b,c) = (1.0,0.67,0.08)$, $r=14$
\item $(a,b,c) = (0.999,0.271,0.587)$, $r=14$
\item $(a,b,c) = (0.87,0.6,0.7)$, $r=14$
\end{compactitem}
we generate 50 realizations of each Kronecker graph.  For each realization, we compute 
a fit using the objective~\eqref{eq:generalfit} with the choices $D_{\text{sq}},N_{F^2}$ 
and using the combination of approaches from the previous section.  
Figure~\ref{fig:kron-fits} shows distribution of fitted parameters to these 50 
samples.  For all four sets 
of parameters, the fitted results closely match the true values, with fairly small 
variation.  

\begin{figure}
\centering
\subfigure[Kronecker $0.99,0.48,0.25$, $r=14$]{\includegraphics[width=0.45\linewidth]{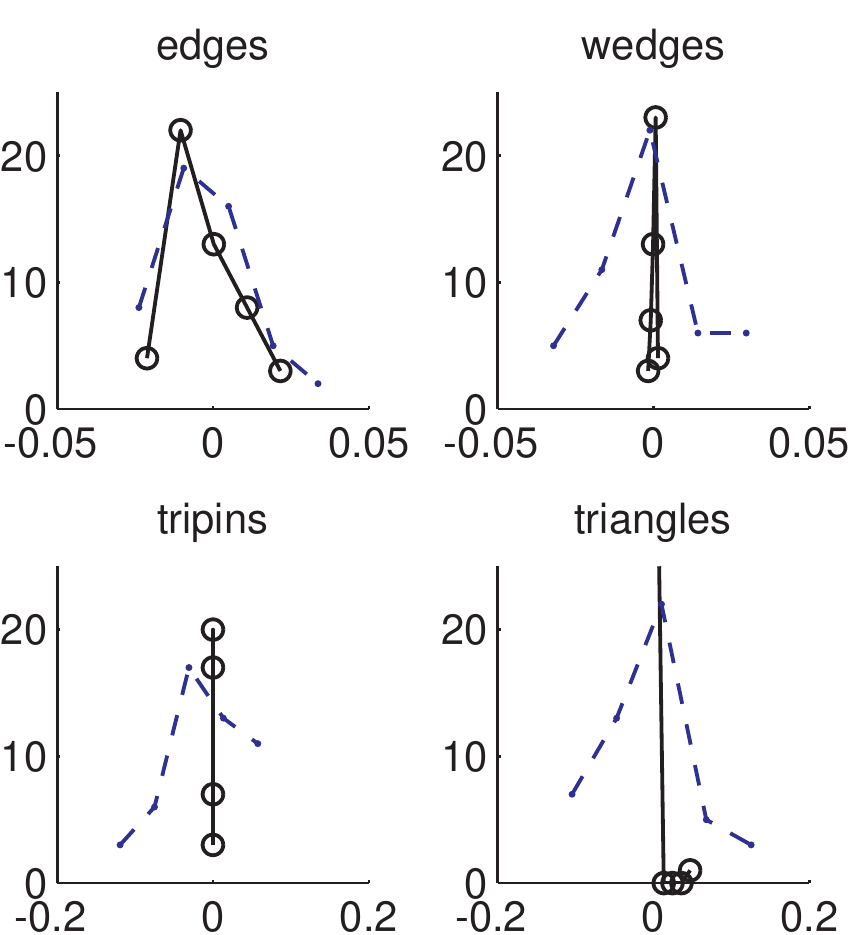}}
\hfil
\subfigure[Kronecker $1.0,0.67,0.08$, $r=14$]{\includegraphics[width=0.45\linewidth]{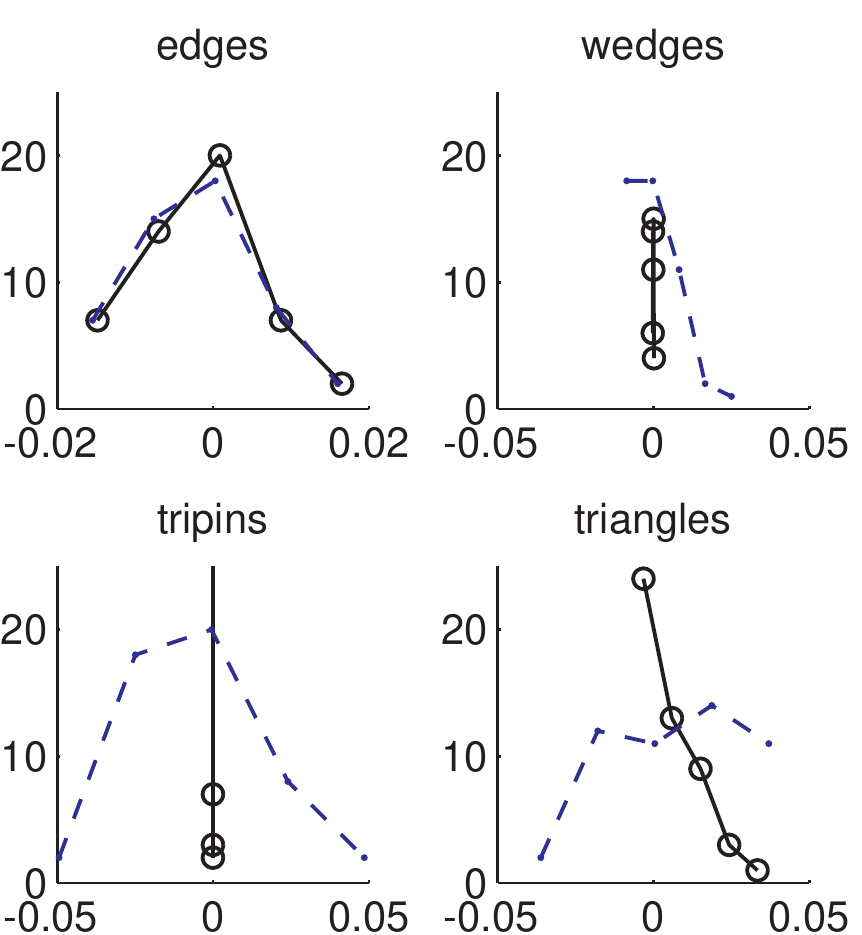}}
\caption{Histograms of the 
relative difference between the true graph feature
and the fitted (black solid) or regenerated (blue dashed line)
feature.  The relative difference is 
$(F_{\text{true}} - F_{\text{fit}})/F_{\text{true}}$.}
\label{fig:kron-feature-errors}
\end{figure}

For these synthetic problems, we also study how the empirical and fitted features 
differ.  Figure~\ref{fig:kron-feature-errors} shows the distribution of the 
relative difference between the expectation of the fitted Kronecker features and the 
actual feature of each realization.  It also shows the difference between the original 
feature count and the feature count of a re-realization.  In other words, generate a 
Kronecker graph, fit the parameters, and re-generate with the fitted parameters.  The 
figures show that the fitted parameters closely match the realizations.  A curious 
property is that the fitted triangle count is always smaller  than the empirical 
count.  The difference in the re-realization can be large, almost 20\% in the case of 
tripins or triangles for the first set of Kronecker parameters.

\begin{figure}
\subfigure[Kronecker $0.99,0.48,0.25$, $r=14$]{\includegraphics[width=0.45\linewidth]{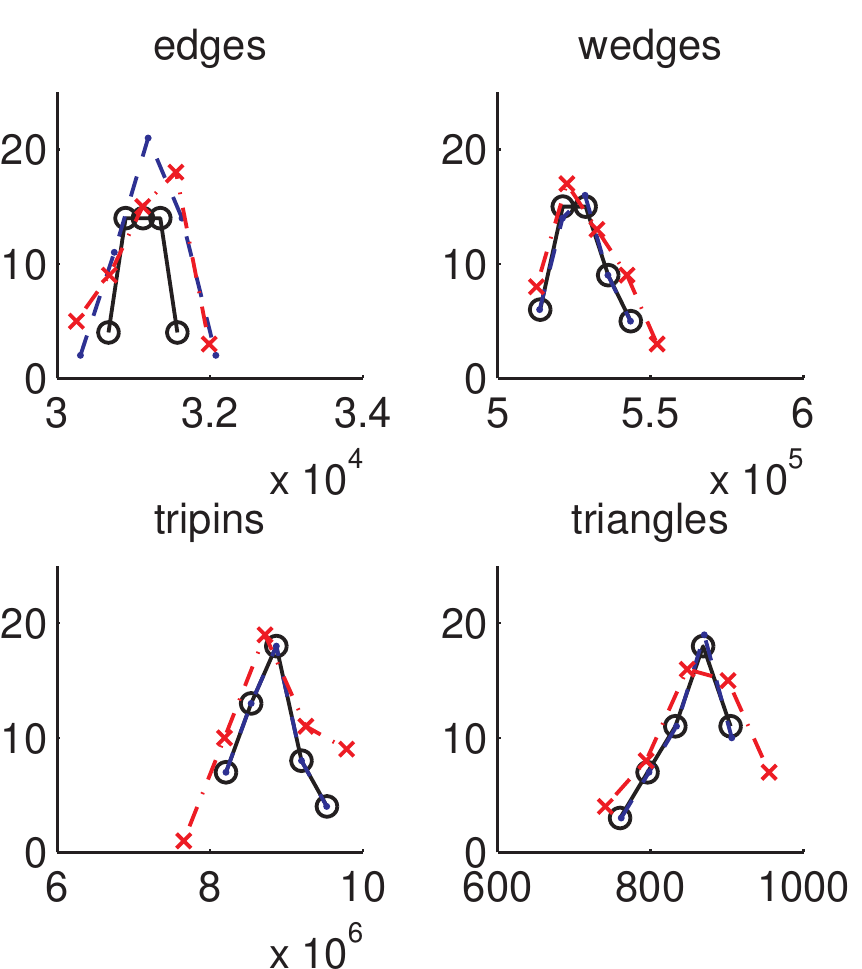}}
\hfil
\subfigure[Kronecker $1.00,0.67,0.08$, $r=14$]{\includegraphics[width=0.45\linewidth]{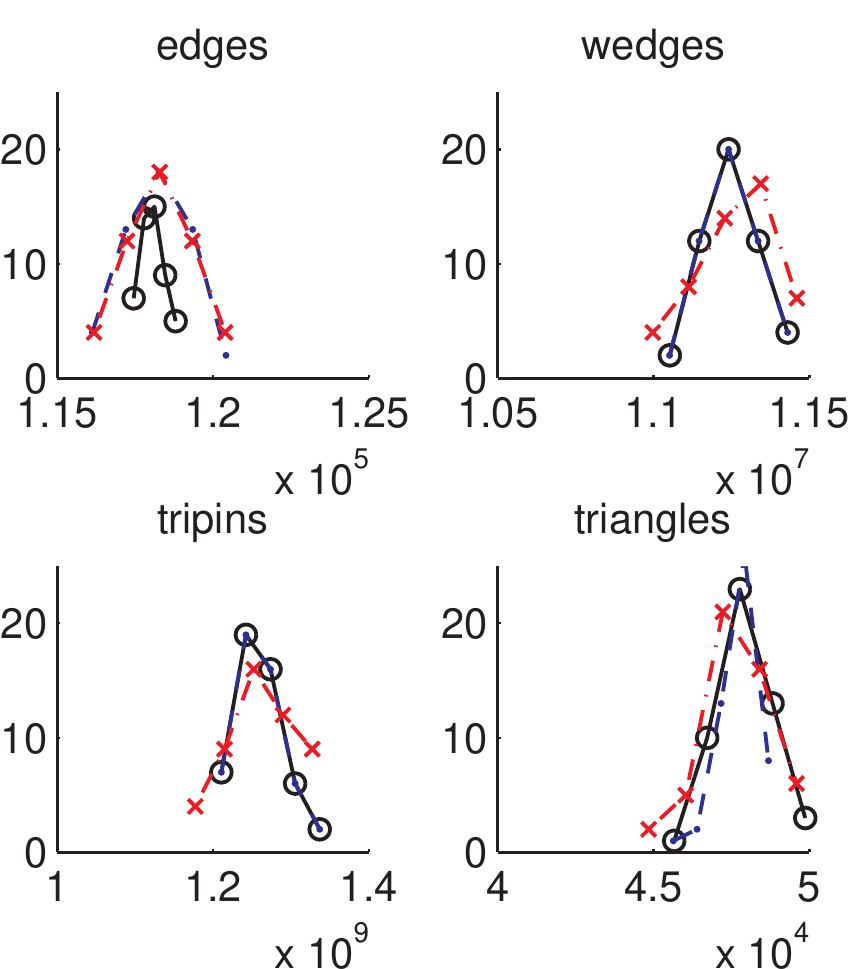}}
\caption{Histograms of empirical features.  The dashed blue line with no $\times$ marks
shows the empirically measured features and 
the solid black line with circles shows the expected value of each feature given the
fitted parameters.  These lines are often on top of each other.
The dashed red line with $\times$ marks shows the features of a regenerated graph.}
\label{fig:kron-features}
\end{figure}

Our final study is the distributions of the graph features given the Kronecker 
parameters, the expected features of the fitted parameters, and the graph features of a 
re-realized Kronecker graph.  The plots in Figure~\ref{fig:kron-features} show that 
these distributions are all quite similar.

\section{Conclusions}\label{sec:conclusions}

We have presented formuals for expected feature counts
in Kronecker graphs and used them to generate
a method of moments fitting strategy.  We found
that summing squared relative feature count errors was robust
and easy to optimize.  For graphs generated by the
Kronecker model, our parameter and feature estimates
closely match those of the fitted graph.  For real world
graphs we often find that the fitted Kronecker model
implies smaller feature counts (apart from edges)
than are seen in the real graph.  The moment estimators
typically come closer to the counts than those from KronFit.

\section*{Acknowledgments}

We thank Tamara Kolda, C. Seshadhri, and Ali Pinar
for helpful discussions.
This work was supported by DMS-0906056
of the National Science Foundation.

\bibliographystyle{plain}
\bibliography{kroneckergraph,gleich}
\end{document}